\documentclass[final]{cvpr}

\usepackage{times}
\usepackage{epsfig}
\usepackage{graphicx}
\usepackage{amsmath}
\usepackage{amssymb}
\usepackage{booktabs}
\usepackage{multirow}
\usepackage{url}

\usepackage[pagebackref=true,breaklinks=true,colorlinks,bookmarks=false]{hyperref}

\def\cvprPaperID{****} 
\def\confYear{CVPR 2023}

\begin{document}

\title{MIPI 2023 Challenge on Nighttime Flare Removal: Methods and Results}

\author{
Yuekun Dai \and Chongyi Li \and Shangchen Zhou \and Ruicheng Feng \and Qingpeng Zhu \and Qianhui Sun \and Wenxiu Sun \and  Chen Change Loy \and Jinwei Gu \and
Shuai Liu \and Hao Wang \and Chaoyu Feng \and Luyang Wang \and Guangqi Shao \and Chenguang Zhang \and Xiaotao Wang \and Lei Lei \and
Dafeng Zhang \and Xiangyu Kong \and Guanqun Liu \and Mengmeng Bai \and Jia Ouyang \and Xiaobing Wang \and Jiahui Yuan \and
Xinpeng Li \and Chengzhi Jiang \and Ting Jiang \and Wenjie Lin \and QiWu \and Mingyan Han \and Jinting Luo \and Lei Yu \and Haoqiang Fan \and Shuaicheng Liu \and
Bo Yan \and Zhuang Li \and Yadong Li \and Hongbin Wang \and
Soonyong Song \and
Minghan Fu \and Rayyan Azam Khan \and Fangxiang Wu \and
Zhao Zhang \and Suiyi Zhao \and Huan Zheng \and Yangcheng Gao \and Yanyan Wei \and Jiahuan Ren \and Bo Wang \and Yan Luo \and
Shuaibo Gao \and Wenhui Wu \and Sicong Kang \and
Nikhil Akalwadi \and Ankit Raichur \and Vinod Patil \and Allabakash G \and Swaroop A \and Amogh Joshi \and Chaitra Desai \and Ramesh Ashok Tabib \and Ujwala Patil \and Uma Mudenagudi
Sicheng Li \and Ruoxi Zhu \and Jiazheng Lian \and Shusong Xu \and Zihao Liu \and
Sabari Nathan \and Priya Kansal
}

\maketitle

\begin{abstract}
\vspace{-4mm}
Developing and integrating advanced image sensors with novel algorithms in camera systems are prevalent with the increasing demand for computational photography and imaging on mobile platforms.
However, the lack of high-quality data for research and the rare opportunity for in-depth exchange of views from industry and academia constrain the development of mobile intelligent photography and imaging (MIPI).
With the success of the \href{http://mipi-challenge.org/}{1st MIPI Workshop@ECCV 2022}, we introduce the second MIPI challenge including four tracks focusing on novel image sensors and imaging algorithms.
In this paper, we summarize and review the Nighttime Flare Removal track on MIPI 2023.
In total, 120 participants were successfully registered, and 11 teams submitted results in the final testing phase.
The developed solutions in this challenge achieved state-of-the-art performance on Nighttime Flare Removal.
A detailed description of all models developed in this challenge is provided in this paper.
More details of this challenge and the link to the dataset can be found at \href{https://mipi-challenge.org/MIPI2023/}{https://mipi-challenge.org/MIPI2023/}.
\end{abstract}

\newcommand{\email}[1]{\href{mailto:#1}{\texttt{#1}}}
{\let\thefootnote\relax\footnotetext{%
\tiny  Yuekun Dai$^{1}$ (\email{ykdai005@ntu.edu.sg}), Chongyi Li$^{1}$ (\email{chongyi.li@ntu.edu.sg}), Shangchen Zhou$^{1}$, Ruicheng Feng$^{1}$, Qingpeng Zhu$^{3}$, Qianhui Sun$^{2}$, Wenxiu Sun$^{3}$, Chen Change Loy$^{1}$, Jinwei Gu$^{2,3}$ are the MIPI 2023 challenge organizers
($^{1}$Nanyang Technological University, $^{2}$SenseBrain, $^{3}$SenseTime Research and Tetras.AI). The other authors participated in the challenge. Please refer to Appendix for details.
\\
MIPI 2023 challenge website: \href{https://mipi-challenge.org/MIPI2023/}{https://mipi-challenge.org/MIPI2023/}
}
}

\section{Introduction}
Lens flare is a common optical phenomenon that occurs when intense light is scattered or reflected within a lens system, resulting in a distinctive radial-shaped bright area and light spots in captured photos. In mobile platforms such as monitor lenses, smartphone cameras, UAVs, and autonomous driving cameras, daily wear and tear, fingerprints, and dust can function as a grating, exacerbating lens flare and making it particularly noticeable at night. Thus, flare removal algorithms are highly desired

Flares can be categorized into three main types: scattering flares, reflective flares, and lens orbs. In this competition, we mainly focus on removing the scattering flares, as they are the most prevalent type of nighttime image degradation. Early attempts at scattering flare removal were made by Wu et al.~\cite{how_to}, who proposed a dataset with physically-based synthetic flares and flare photos taken in a darkroom. However, this approach did not perform well in nighttime situation. To address this issue, Dai et al.~\cite{dai2022flare7k,dai2023nighttime} created a new synthetic dataset specifically designed for nighttime scenes. This challenge is based on the subset of \cite{dai2022flare7k} and aims to restore the flares-corrupted images with different complicated degradations. Further details will be discussed in the following sections.

We hold this challenge in conjunction with the second MIPI Challenge which will be held on CVPR 2023. Similar to the first MIPI challenge~\cite{feng2023mipi,sun2023mipi,yang2023mipi,yang2023mipi2,yang2023mipi3}, we are seeking an efficient and high-performance image restoration algorithm to be used for recovering flare corrupted images. MIPI 2023 consists of four competition tracks:

\begin{itemize}
    \item \textbf{RGB+ToF Depth Completion} uses sparse and noisy ToF depth measurements with RGB images to obtain a complete depth map.
    \item \textbf{RGBW Sensor Fusion} fuses Bayer data and a monochrome channel data into Bayer format to increase SNR and spatial resolution.
    \item \textbf{RGBW Sensor Re-mosaic} converts RGBW RAW data into Bayer format so that it can be processed by standard ISPs.
    \item \textbf{Nighttime Flare Removal} is to improve nighttime image
quality by removing lens flare effects.
\end{itemize}

\section{MIPI 2023 Nighttime Flare Removal}
To facilitate the development of efficient and high-performance flare removal solutions, we provide a high-quality dataset to be used for training and testing and a set of evaluation metrics that can measure the performance of developed solutions.
This challenge aims to advance research on nighttime flare removal.

\subsection{Datasets}
In this competition, we will provide a synthetic flare dataset with 5,000 scattering flare images~\cite{dai2022flare7k} in 1440$\times$1440$\times$3. These images can be added to the provided flare-free background to synthesize paired data for training. We will provide detailed annotations for each component of the synthetic flare, including streak, glare, and light source. To help the participants evaluate the performance of their models, we provide 100 real captured flare-corrupted/flare-free validation image pairs. Finally, models will be evaluated on 100 extra captured test image pairs.

\subsection{Evaluation}

In this competition, we use the standard Peak Signal To Noise Ratio (PSNR) as our evaluation metrics. The restoration results in the glare and streak region will be evaluated by using glare region PSNR (G-PSNR) and streak region PSNR (S-PSNR). Since the ground truth images can not offer perfect light source image without any flares, we only compute the PSNR for the region without the light source in this challange. Finally, global PSNR is calculated in the region without light source. These three metrics will be averaged to get the final score. Participants can view their score, G-PSNR, S-PSNR of their submission to optimize the model's performance on scattering flares' different components.

\subsection{Challenge Phase}
The challenge consisted of the following phases:
\begin{enumerate}
    \item Development: The registered participants get access to the data and baseline code, and are able to train the models and evaluate their running time locally.
    \item Validation: The participants can upload their models to the remote server to check the fidelity scores on the validation dataset, and to compare their results on the validation leaderboard.
    \item Testing: The participants submit their final results, code, models, and factsheets.
\end{enumerate}

\section{Challenge Results}
Among $120$ registered participants, $11$ teams successfully submitted their results, code, and factsheets in the final test phase.
Table \ref{tab:result} reports the final test results and rankings of the teams. 
Only three teams train their models with extra data, and several top-ranked teams apply ensemble strategies (self-ensemble, model ensemble, or both).
The methods evaluated in Table \ref{tab:result} are briefly
described in Section \ref{sec:methods} and the team members are listed in Appendix.

Finally, the MiAlgo team is the first place winner of this challenge, while  Samsung Research China - Beijing (SRC-B) team win the second place and MegFR team is the third place, respectively.
Because the quantitative metrics of the competition can not perfectly reflect everyone's results, we also conducted an additional user study with 20 people. In this user study, we choose all test results for the teams in top 5 and complete the light source for these results. We ask the user to pick the best flare-removed images among top 5 results. Based on the result of user study, we present the \textbf{Best Visualization Award} to the team of Samsung Research China - Beijing.

\begin{table*}
\centering
\caption{Results of MIPI 2023 challenge on nighttime flare removal. ‘Runtime’ for per image is tested and averaged across the validation datasets, and the image size is $512\times512$. ‘Params’ denotes the total number of learnable parameters.}
\label{tab:result}
\scalebox{0.7}{
\begin{tabular}{ll|ccc|ccccc}
\toprule
\multicolumn{1}{c}{}                            & \multicolumn{1}{c}{}                            & \multicolumn{3}{|c|}{Metric}                                                               & \multicolumn{1}{l}{}                             &                               &                            &                              &                            \\
\multicolumn{1}{c}{\multirow{-2}{*}{Team Name}} & \multicolumn{1}{c|}{\multirow{-2}{*}{User Name}} & Score & S-PSNR & G-PSNR & \multicolumn{1}{l}{\multirow{-2}{*}{Params (M)}} & \multirow{-2}{*}{Runtime (s)} & \multirow{-2}{*}{Platform} & \multirow{-2}{*}{Extra data} & \multirow{-2}{*}{Ensemble} \\
\midrule
MiAlgo             & mialgo\_ls               & $29.44 _{(1)}$  & $28.59_{(1)}$           & $28.89_{(1)}$        & 65.56             & 2.2      & Nvidia RTX 3090                & Yes          & self-ensemble + model                      \\
SRC-B            & xiaozhazha                    & $29.16_{(2)}$  & $28.21_{(2)}$           & $28.59_{(2)}$        & 23.56             & 2.0      & Nvidia A100               & Yes            & self-ensemble + model                         \\
MegFR                   & jiangchengzhi, lxp0\_0              & $27.66_{(3)}$  & $26.56_{(3)}$           & $27.05_{(3)}$        & 20.4             & 0.168      & Nvidia 2080Ti                   & -            & self-ensemble      \\
AntIns                 & Yjingyu              & $26.03_{(4)}$  & $24.04_{(6)}$          & $25.68_{(5)}$       & 50.88              & 2.83         & Nvidia Tesla V100                 & -            & -                            \\
ActionBrain-ETRI           & soony               & $26.03_{(5)}$  & $24.49_{(4)}$           & $25.36_{(8)}$        & 41              & 0.24      & Nvidia RTX A6000                 & -            & -                          \\
USask-Flare                   & eason           & $25.86_{(6)}$  & $24.19_{(5)}$           & $25.38_{(6)}$        & 5.3             & 0.64      & Nvidia RTX 3090                   & -            & -                          \\
LVGroup\_HFUT                    & HuanZheng                     & $25.82_{(7)}$  & $23.36_{(9)}$           & $25.92_{(4)}$        &     29.16              &  0.03         & Nvidia RTX 3090                   &   -          & -              \\
szzzzz01             & gsb                    & $25.72_{(8)}$  & $24.00_{(7)}$           & $25.26_{(9)}$        & /              & 0.47      & Nvidia A100                 & -            & -             \\
CEVI\_Explorers                  & Lowlight\_Hypnotise,Niksx                  & $25.52_{(9)}$  & $23.40_{(8)}$           & $25.26_{(10)}$        & 82.2              & 2.07      & Nvidia RTX 3090                  & -            & -              \\
AI ISP              & zrx              & $25.46_{(10)}$ & $23.17_{(10)}$           & $25.38_{(7)}$        & 21.71             & 0.29      & Nvidia A100                 & Yes            & -              \\
Couger AI          & SabariNathan          & $20.76_{(11)}$ & $15.88_{(11)}$          & $22.71_{(11)}$       &    1.6428      & 0.21          & Nvidia Tesla V100               &       -       &          -                 \\
\bottomrule
\end{tabular}
}
\end{table*}

\section{Methods}\label{sec:methods}

\begin{figure*}[htbp]
    \vspace{-2mm}
    \centering
    \includegraphics[width=1\textwidth]{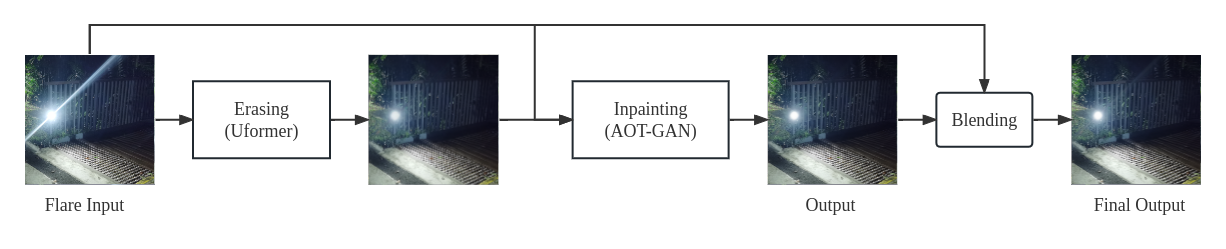}
    \vspace{-11mm}
    \caption{The network architecture of MiAlgo team.}
    \vspace{-3mm}
    \label{fig:mialgo}
\end{figure*}

\paragraph{\bf MiAlgo.}
For the nighttime flare removal task, this team proposes a two-stage network(see Figure~\ref{fig:mialgo}) that consists of an erasing module and an inpainting module. In the first stage, borrowing from \cite{dai2022flare7k}, they use the Uformer\cite{Uformer} as the erasing module, aiming to remove the flares as much as possible. In the second stage, they use the AOT-GAN\cite{zeng2022aggregated} as the inpainting module to restore the details on the flare area, the input is the original flare image concatenated with the output of erasing module. The output of the inpainting module has excellent visualization effects, but its PSNR is relatively low. They figure out that the ground-truth image is not entirely flare-free, so they blend the output of the inpainting module with the original flare input to get the final output, which improves the PSNR by 0.4 dB. Regarding data generation, they collect several nighttime images as base images and randomly adjust the colors of the flares to blue, yellow, and white, which are common on the ground. They also add light and local haze to the base image to simulate the more realistic flare-corrupted images. The data augmentation uses very common methods such as random affine, color jitter, etc. 

When training, they first train the erasing module using synthetic data generated online for about 100,000 iterations. Then fix the weight of the erasing module and train the inpainting module for about 100,000 iterations. The initial learning rate is 1e-3 and they use cosine annealing to reduce the learning rate. The batch size is 16 and the patch size is 512. They train on 8 Tesla V100 GPUs for about 2 days. During testing, they use the self-ensemble(X8) and the model-ensemble(X4) strategy to get the highest PSNR.

\begin{figure}[t]
	\centering
	\includegraphics[width=1.0\linewidth]{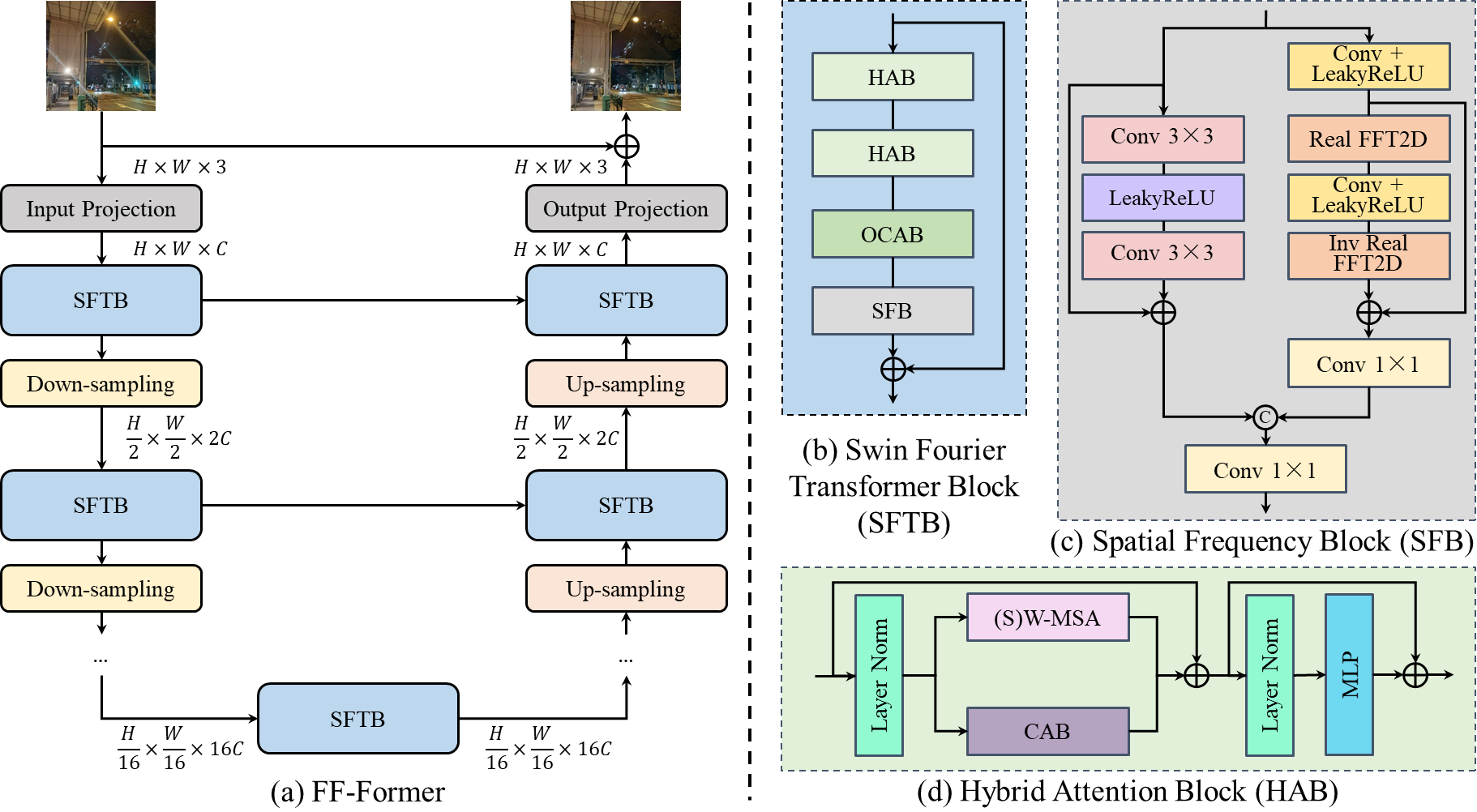}
	\caption{The network architecture of SRC-B team.}
	\label{fig:FF-Former}
    \vspace{-3mm}
\end{figure}

\paragraph{\bf SRC-B.}
In the process of removing nighttime flare, it is crucial to have a large receptive field because flare can occupy a substantial portion of an image, even potentially the entire image. However, the conventional window-based Transformer approaches restrict the receptive field within the window, limiting its ability to capture global features. And the flare can cause the dark regions to become brighter and result in a loss of contrast and alteration of the frequency characteristics of the image. To address these challenges, SRC-B introduced FF-Former (see Figure~\ref{fig:FF-Former}), which is based on Fast Fourier Convolution (FFC) and is designed to extract global frequency features for enhancing nighttime flare removal. To achieve this, SRC-B incorporates a Spatial Frequency Block (SFB)~\cite{zhang2022swinfir} after the Swin Transformer, which forms the Swin Fourier Transformer Block (SFTB). This configuration enables the establishment of long dependencies and the extraction of global features. Unlike the traditional Transformer, which relies on global self-attention, the SFB module only performs convolution computation, making it both effective and efficient. 

SFTB is the basic module for our FF-Former, and the channel number $C$ is 32 in the first SFTB of Encoder. We set 4 levels in the Encoder and Decoder module for extracting multi-scale features. Following Dai et al.~\cite{dai2022flare7k}, we crop the input flare-free and flare-corrupted images into 512$\times$512 with batch size of 2 to train our FF-Former. We use the Adam with  $\beta_1=0.9$ and $\beta_2=0.99$ to optimize the Charbonnier L1 loss function~\cite{lai2018fast} for nighttime flare removal. The initial learning rate is 1e-4 and we use CosineAnnealingLR with 600,000 maximum iterations and 1e-7 minimum learning rate to adjust the learning rate. we also use horizontal and vertical flips for data enhancement.

\paragraph{\bf MegFR.}
The team MegFR proposes a new data synthesis method that can synthesize flare images with distributions that are closer to real data and propose a new learning method based on signal separation to remove flare more finely.

Since it is difficult to obtain real Flare-corrupted images and flare-free image pairs, this team uses the officially provided flare images and background images to synthesize a large number of image pairs for training. Given a flare-free image \(I_b\) as background, they first randomly select \(t\) from \(5000\) flare images and denote it as \(F=\{f_1,\dots,f_t\}\), in this paper \(t\) is set to 4. Firstly, they perform random inverse gamma correction on \(I_b\) and \(f_i\):
\begin{equation}
\begin{aligned}
I_b^g=\gamma^{-1}(I_b;\theta^y),\\
f_i^g=\gamma^{-1}(f_i;\theta_i^f),
\end{aligned}
\end{equation}
where \(\theta^y\) and \(\theta_i^f\) are random numbers in \([1.8,2.2]\), which represents the correction strength. Then in order to simulate more morphological flares, they do a Gaussian blur with random intensity on each flare:
\begin{equation}
\begin{aligned}
f_i^b=g(f_i^g;\theta_i^b),
\end{aligned}
\end{equation}
where \(g\) represents Gaussian blur, and \(\theta_i^b\) represents the size of the Gaussian kernel, which is a random number in \([5,21]\). At the same time, in order to simulate the different brightness of the main light source and other light sources in the same scene, this team multiplies different gain values for different flares. Then after randomly offsetting the positions of all the flares, finally these flares are added to get the combination of multiple flares.
\begin{equation}
\begin{aligned}
f_c=\sum_{i=1}^t gain_i*\Phi(f_i^b),
\end{aligned}
\end{equation}
where \(\Phi\) means randomly offsetting the center of the flares, and \(gain_i\) is a random number in \([0.8,1.0]\). Finally, this team gets the flare-corrupted image \(I_x\) through:
\begin{equation}
\begin{aligned}
I_x= \gamma(I_b^g+f_c;\theta),
\end{aligned}
\end{equation}
where \(\gamma\) is gamma correction, \(\theta\) is a random number in \([1.8,2.2]\). The corresponding flare-free image \(I_y\) and flare image \(I_f\) could be obtained in the following:
\begin{equation}
\begin{aligned}
I_y= \gamma(I_b^g;\theta),\\
I_f= \gamma(f_c;\theta).
\end{aligned}
\end{equation}

\begin{figure}[t]
	\centering
	\includegraphics[width=0.45\textwidth]{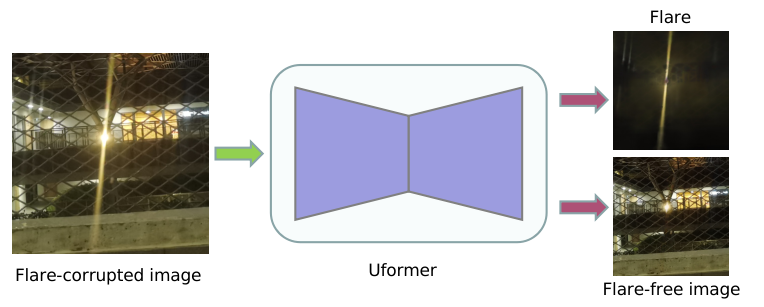}
	\caption{The pipeline of MegFR team.}
	\label{MefFR}
	\vspace{-0.5cm}
\end{figure}

As shown in Fig.~\ref{MefFR},after generating the data, this team uses a Uformer~\cite{wang2022uformer} \(\psi\) for signal separation:
\begin{equation}
\begin{aligned}
Y=\psi(I_x),
\end{aligned}
\end{equation}
where \(Y\in\mathbb{R}^{w\times{h}\times{6}}\) is the result of signal separation, the first three channels \(Y_c\) are image content signals, and the last three channels \(Y_f\) are flare signals. This team uses L1 loss to supervise and train the \(\psi\):
\begin{equation}
\label{eq:loss}
\begin{aligned}
loss_{L1} = \lvert Y_c-I_y\rvert+\lvert Y_f-I_f\rvert.
\end{aligned}
\end{equation}

Perceptual loss can facilitate regularizing the network to produce images having more structural similarity with the ground truth ~\cite{johnson2016perceptual,sr}. So this team further introduces perceptual loss into our training: 
\begin{equation}
loss_p= \lvert f_p-f_g\rvert^2,
\end{equation}
where \(f_p\) and \(f_g\) respectively represent the features of \(Y_c\) and \(I_y\). In particular, the features here refer to the features of the last convolution of an alexNet pre-trained on ImageNet.

The final loss is a combination of the two losses:
\begin{equation}
\begin{aligned}
loss=loss_{L1}+loss_p.
\end{aligned}
\end{equation}

\vspace{-3mm}
\paragraph{\bf AntIns.}
The team AntIns proposes an effective nighttime flare removal pipeline for MIPI challenge. Firstly, they employed a strong image restoration model, Uformer\cite{Uformer}, as a base model. Then, some improvements were made to the loss function, and an effective data augmentation method was proposed for the problem of the nighttime flare removal. Finally, through model fusion and test-time augmentation, the performance of the model was further improved and ultimately achieved competitive results in the challenge.

\begin{figure}[t]
	\centering
	\includegraphics[width=0.45\textwidth]{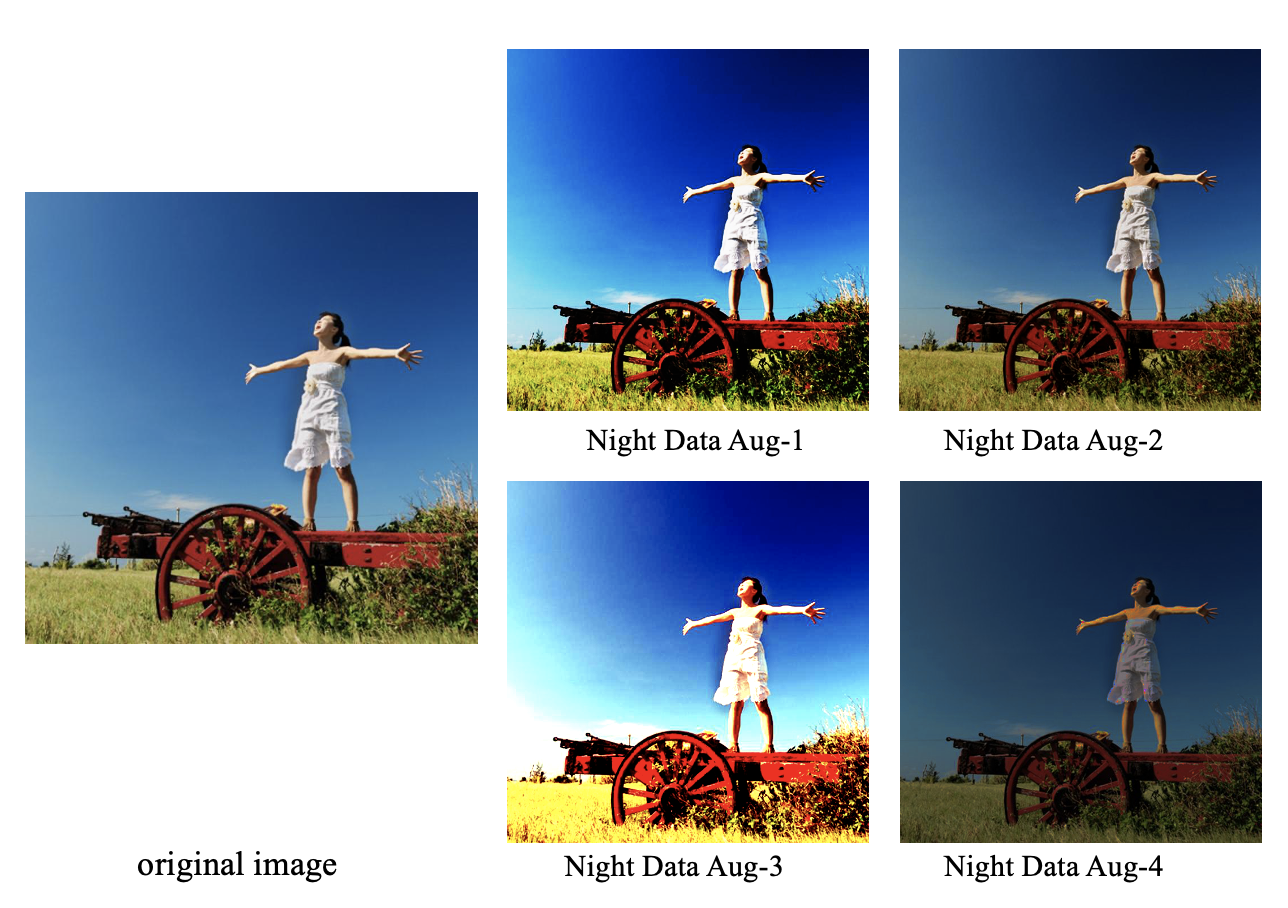}
	\caption{Augmented image examples of Night Data Aug strategy.}
	\label{ant1}
	\vspace{-0.4cm}
\end{figure}

The team uses Uformer-B(Base) as a base model, which is trained from scratch. They synthesize training images using Flickr24K as background images and 5k scattering flare images from Flare7K. The team also designs a night data augmentation strategy (Night Data Aug) for the background images. As shown in Figure \ref{ant1}, the Night Data Aug has four modes, with one randomly selected for each image during training. The loss function includes both L1 loss and perceptual loss. Additionally, the team assign different weights to the L1 loss for areas inside and outside the flare, and the weight for areas inside the flare is 5, while outside the flare it is 1. The weight of perceptual loss is 1. The training samples use a patch size of 512, and the optimizer is AdamW~\cite{loshchilov2018decoupled} with an initial learning rate of 0.0002. The team trains for 85 epochs and find that longer training might improve the results. 
Ultimately, the proposed pipeline achieves a score of 26.03 on the test set of the MIPI challenge.


\paragraph{\bf ActionBrain-ETRI.}

\begin{figure}[htbp]
  \centering
  \includegraphics[width=0.5\textwidth]{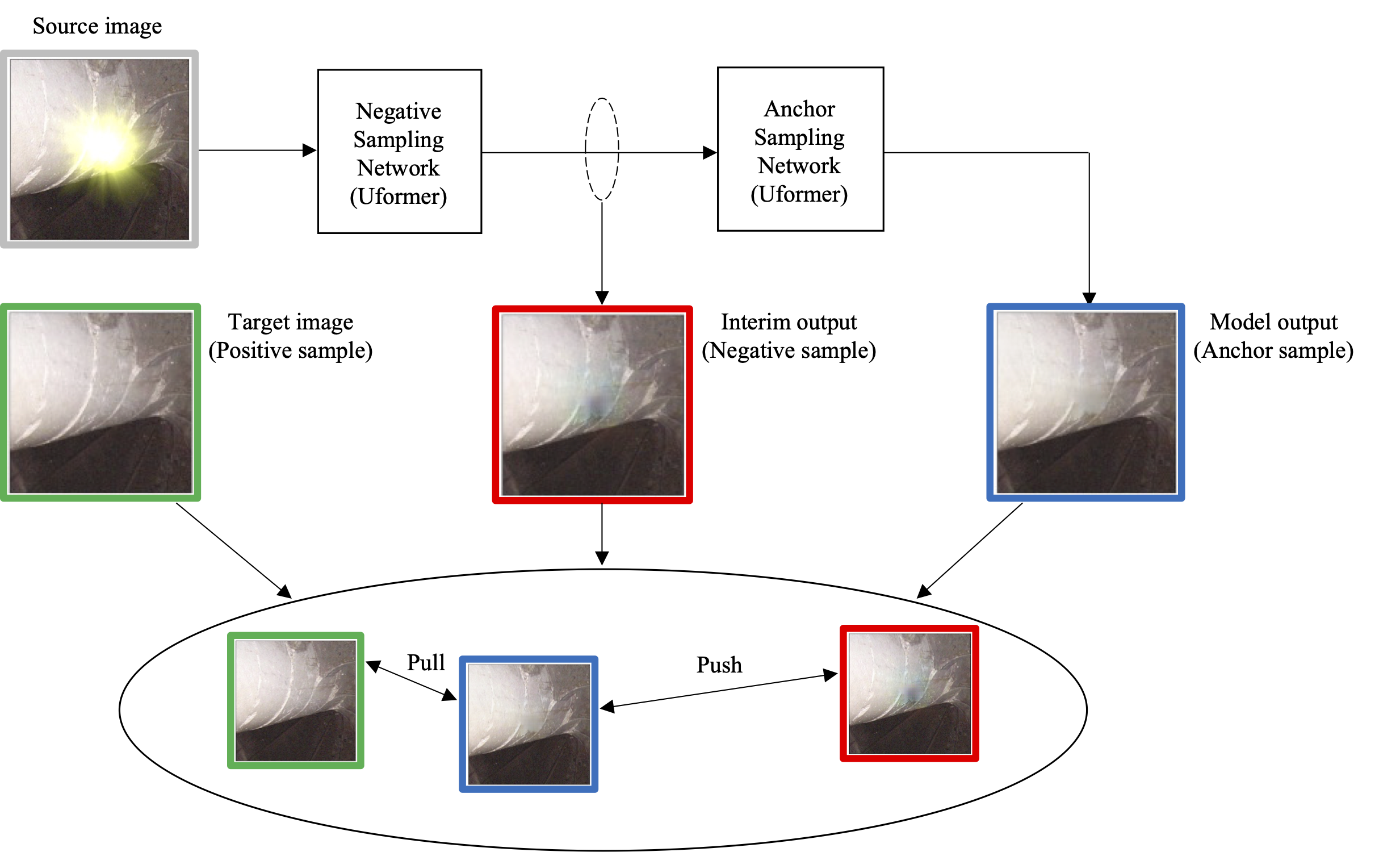}
  \caption{The network architecture of ActionBrain-ETRI team.}
  \label{pipeline_actionbrain_etri}
\end{figure}

This team proposes a cascaded neural network to reduce distortions by flares in dark images.
This network is designed to concatenate the identical Uformer models serially as shown in Figure \ref{pipeline_actionbrain_etri}.
The proposed network assums that the second part of the model is able to compensate for errors induced from the first one \cite{wu2021contrastive}.
The Uformer model is provided as one of the baseline codes.
The input shape of the model is $batchsize \times 3 \times 512 \times 512$, and the output shape is $batchsize \times 6 \times 512 \times 512$.
The earlier 3 channels in the output shape are image predictions, and the rest 3 channels are mask predictions.
This team uses image predictions only.
Weights of the first model are configured by copying provided pre-trained checkpoint.
Their gradients are fixed to get stable negative samples.
The second model is set to be trainable, and it produced anchor samples.
This team assumes that target images were dealt with positive samples.
The proposed network is optimized by MSE $L_{MSE}$, triplet \cite{schroff2015facenet} $L_{triplet}$ and perceptual loss \cite{sr} $L_{PL}$:
\begin{align} 
\begin{split}
& L_{ActionBrain-ETRI}(x^a_i, x^p_i, x^n_i) \\ 
& = \lambda \times L_{MSE}(x^a_i, x^p_i) + \delta \times L_{PL}(x^a_i, x^p_i) \\
& + (1-\lambda) \times L_{triplet}(x^a_i, x^p_i, x^n_i), 
\end{split}
\end{align}
where $x_i^n$, $x_i^p$, $x_i^a$ are negative, positive, anchor samples of the $i$-th triplet, $\lambda=0.6$, and $\delta=0.001$.

To train the proposed network, dual GPUs (NVIDIA Quadro RTX A6000) are used.
Then the models are trained using provided images and pre-trained checkpoints without extra resources.
The proposed network has 40,892,876 parameters (20,446,438 trainable and 20,446,438 non-trainable parameters).
Here, hyperparameters for learning rate, batch size, number of workers, and epochs are set to 0.0002, 6, 6, and 50 respectively.
At each training and testing step, time is spent around 2900 and 24 seconds per epoch.
Based on calculations, the inference time per image is estimated to be 0.24 seconds, given that the number of target images is 100.

\begin{figure*}[htbp]
	\centering
	\includegraphics[width=0.8\linewidth]{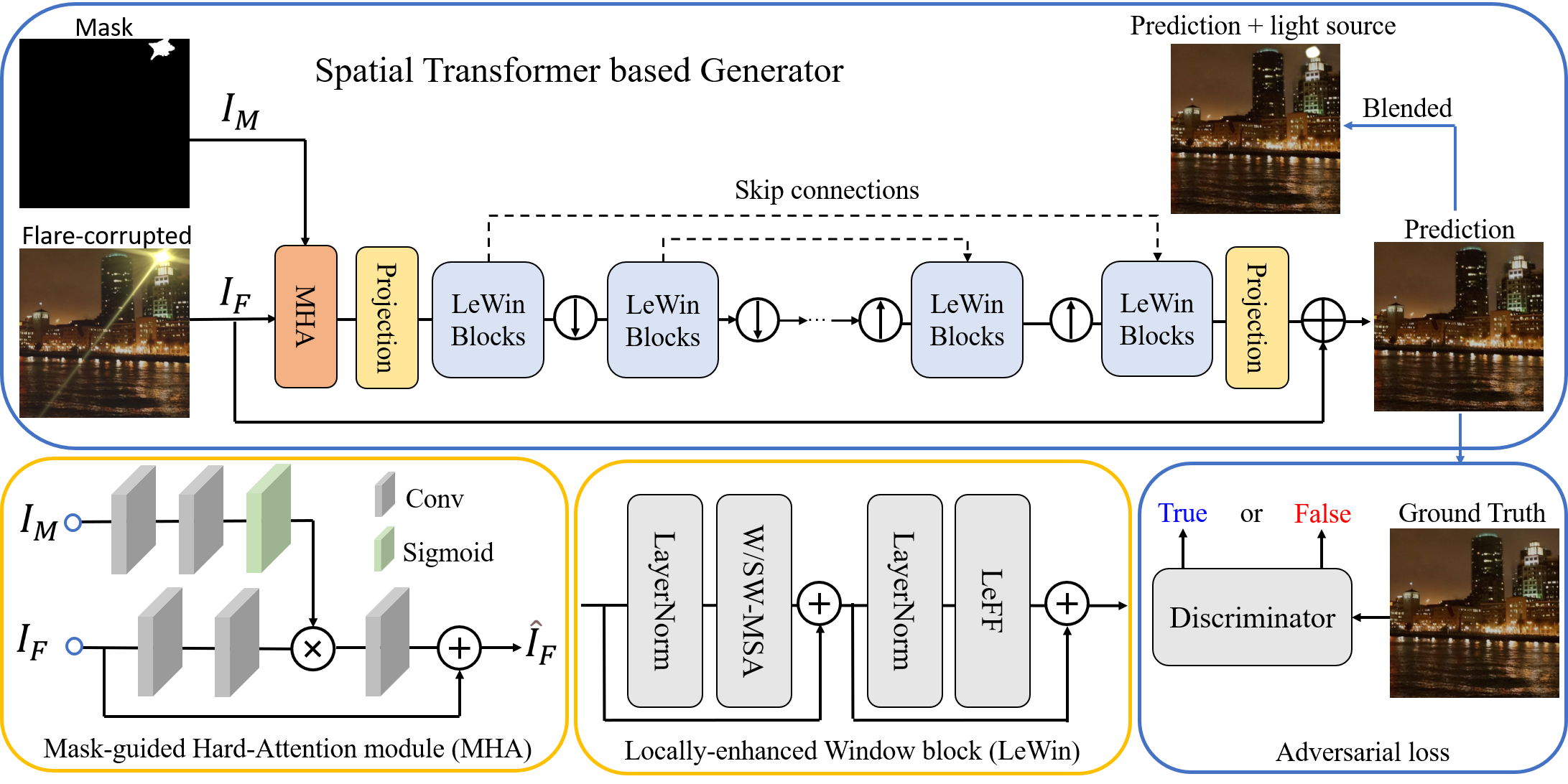}
	\caption{The network architecture of USask-Flare team.}
    \vspace{-4mm}
	\label{pipeline_USask_Flare}
\end{figure*}

\paragraph{\bf USask-Flare.} This team proposes a novel Light source guided Spatial transformer Generative Adversarial Network, namely LS-GAN. The overall network pipeline is shown in Figure~\ref{pipeline_USask_Flare}. LS-GAN follows a U-Shaped structure, which has an encoder, a decoder and massive skip connections. The Locally-enhanced Window (LeWin) block is adopting the design in Uformer \cite{Uformer} and SRGAN \cite{sr} is adopted as the discriminator to form the adversarial loss $L_{adv}$. LS-GAN has two key components to achieve better flare-removal performance, which contain the designed Mask-guided Hard-Attention module (MHA module) and the hybrid objective function (including smooth L1 loss, 
perceptual loss, gradient loss, and adversarial loss). 

They observe that the conditions of the light source in the image play an important role in the resulting flares. Thus, in the designed MHA module, this team uses light source masks as strong prior knowledge to lead the network training. In detail, they use the image saturation threshold and use image opening operation to preprocess the network input flare images and localize the light source area. By using the sigmoid function as the activation function, the light source areas get higher initial weights and non-light source areas get initial lower weights, which allows the network to focus more on the light source zones in the flare reflection removal process. Their proposed hybrid loss can be defined as:
\begin{equation}
\begin{aligned}
\mathcal{L}_{\text {total }}=\mathcal{L}_{\text {smooth}}+\alpha \mathcal{L}_{p e r}+ \beta \mathcal{L}_{\text {mge}}+\gamma \mathcal{L}_{a d v},
\end{aligned}
\end{equation}
where $\alpha=0.01, \beta=0.01$ and $\gamma=0.005$ are the hyperparameters weighting for each loss function. They use ImageNet \cite{imagenet} pre-trained VGG16 \cite{vgg} as the loss network to measure perceptual similarity to form the perceptual loss. They use the Sobel operator for gradient calculation on both X-axis and Y-axis to form the mean gradient error (mge) loss. The smooth-L1 loss and adversarial loss are the same as \cite{dw,two_branch}.

To preserve the original light source, this team blends the input light source back into the network prediction images. In detail, they compute a binary mask according to the saturated threshold. This binary mask indicates the light source zone of the input image. The masked area in the network output is then replaced by the input pixels, producing a more realistic final result.

For the network training details, this team resized raw training samples to 512$\times$512 before feeding them into the network. They randomly rotate each image by 90,180,270 degrees and also utilize horizontal and vertical flips as our data augmentation strategies. The batch size is set to 2 due to memory limitation. They use Adam with $\beta_{1}$=0.9, $\beta_{2}$=0.999 as our optimizer, and a specific decay strategy is adopted, where the initial learning rate is set to 1e-4 and decays 0.5 times at 100, and 150 epochs for a total of 500 epochs. The discriminator uses the same optimizer and weight decay strategies.

\paragraph{\bf LVGroup\_HFUT.}

\begin{figure}[t]
	\centering
	\includegraphics[width=0.45\textwidth]{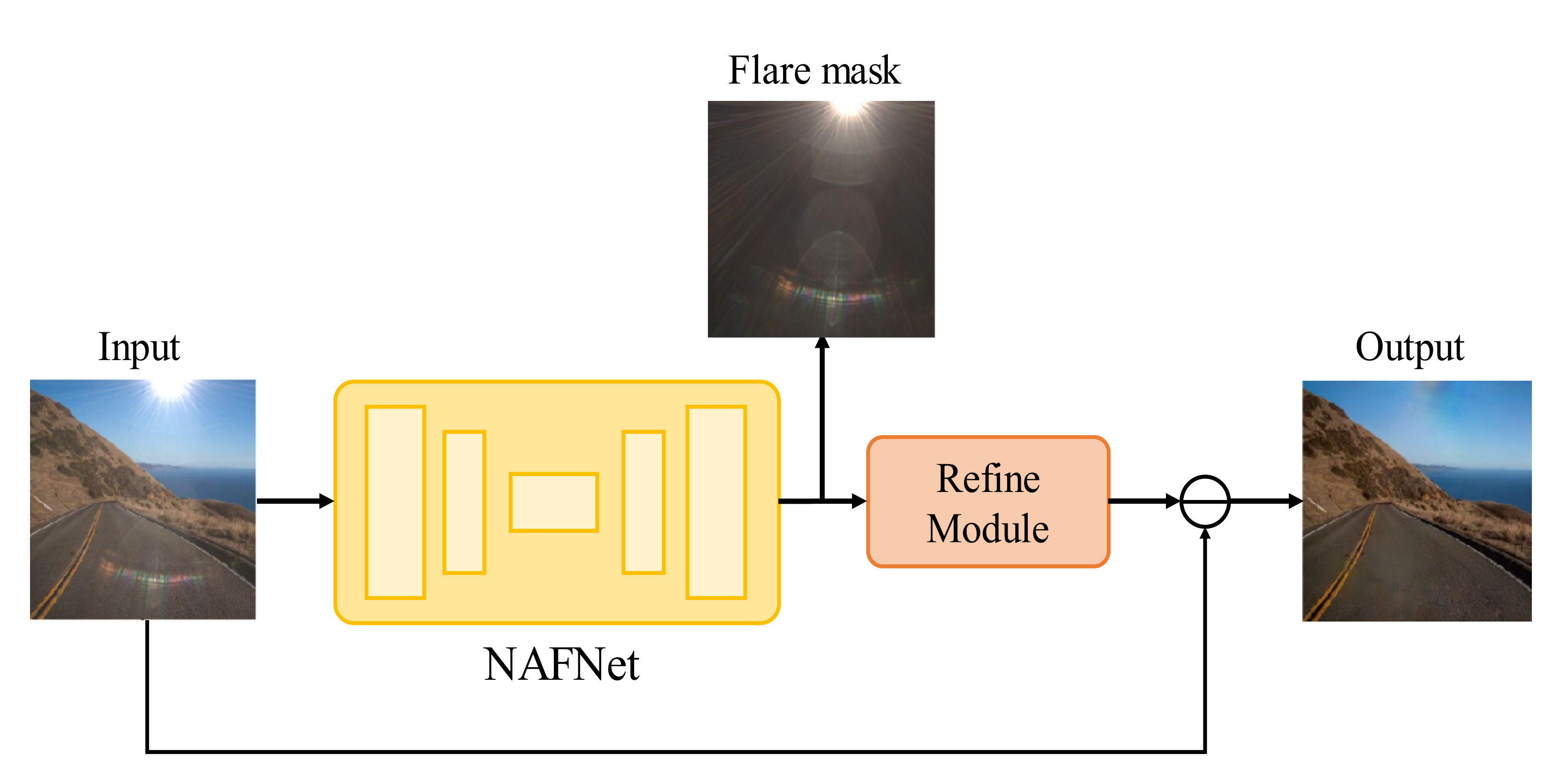}
	\caption{The network architecture of LVGroup\_HFUT team.}
	\label{fig:LVGroup_HFUT}
	\vspace{-0.5cm}
\end{figure}

This team try to build an effective model with flare guidance for nighttime flare removal (see Figure~\ref{fig:LVGroup_HFUT}).
To be specific, the author first employ NAFNet\cite{chen2022simple} as the backbone to generate a flare since NAFNet is a very strong model for image restoration. According to the obtained flare and the synthesis mechanism of flare-interrupted image, the author try to get a flare-free image by subtracting the original image and the flare images. Please note that the authors use the designed refine module to allow for flexible subtraction rather than strict subtraction. The authors use $L_1$ loss and frequency reconstruction loss (weight:0.1) to constrain the predicted flare mask and final flare-free output.
The proposed solution is implemented based on PyTorch version 1.10 and NVIDIA RTX 3090 with 24G memory. During training, the authors first perform a series of data augment operations sequentially as follows: 1):random crop to 384x384; 2): vertical flip with probability 0.5; 3): horizontal flip with probability 0.5. The authors train the model for 90 epochs on provided training dataset with initial learning rate 1e-4 and batch size 6.
For testing (inferring), The authors feed original resolution (512x512) images into the pretrained model and directly generate the final flare-free images.

\begin{figure}[htp]
\centering
\includegraphics[width=1.0\linewidth]{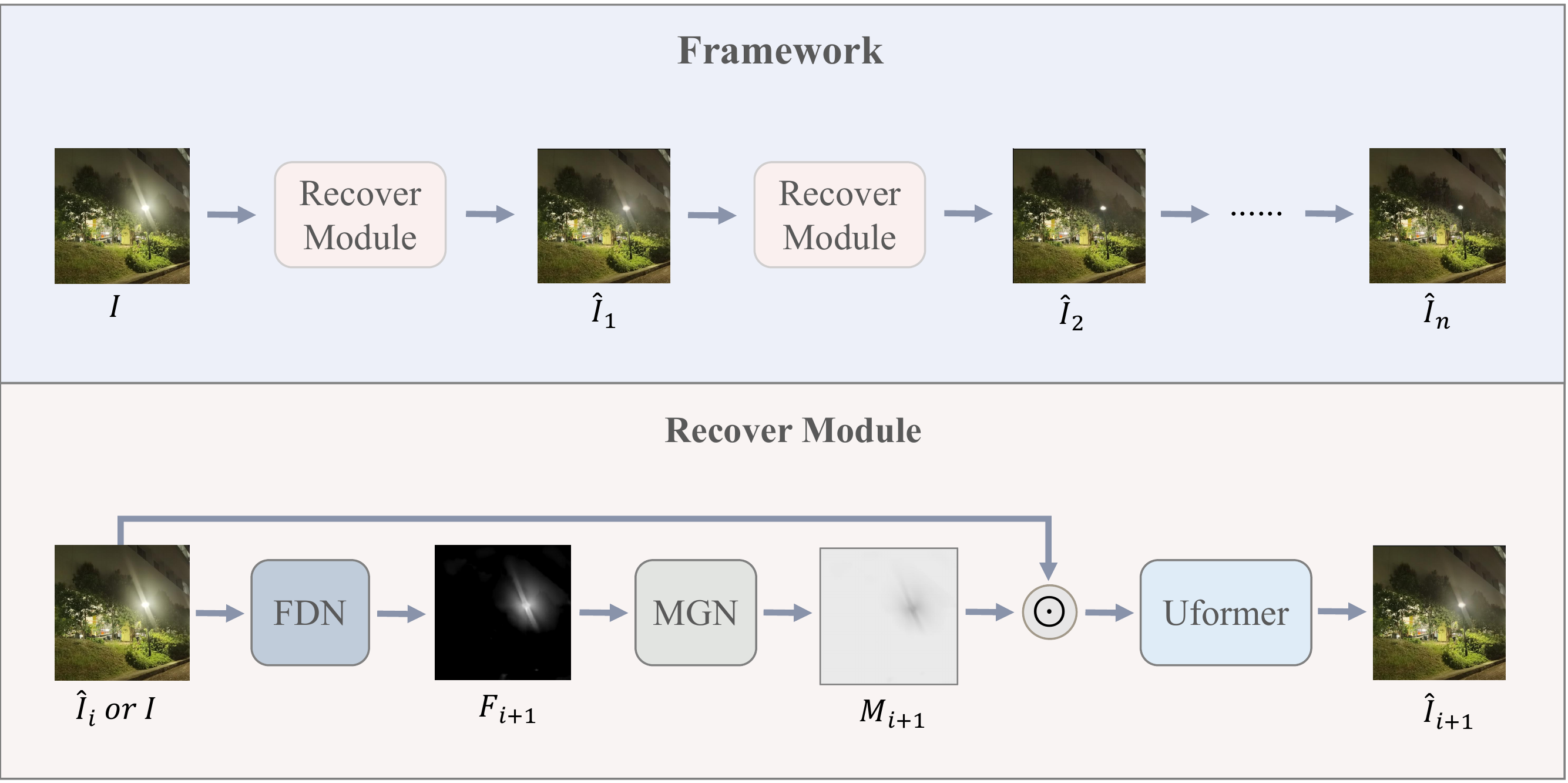}
\caption{The framework of szzzzz01 team.}
\label{fig:szzzzz01}
\vspace{-0.3cm}
\end{figure}

\paragraph{\bf szzzzz01.}This team proposes a flare-region attention recurrent network for flare removal, and the pipeline is shown in Figure. \ref{fig:szzzzz01}. Fundamentally, flare removal is achieved by utilizing surrounding pixels to restore the center corrupted pixel. Therefore, different surrounding corrupted pixels should make a different contribution to the restoration. Based on that, this team designs a spatial attention mask $M$ for a corrupted image before feeding it into the restoration module. To make full use of the attention mask, rather than one-step restoration, they gradually remove the flare within multi-steps. Concretely, a flare detected network (FDN) with a Unet \cite{unet} architecture is first pretrained to generate the rough flare region $F$, whose loss is:
\begin{equation}
\begin{aligned}
\mathcal{L}_{FDN} &= \Vert F - F_{ref}\Vert _2^2 + \Vert \nabla F - \nabla F_{ref}\Vert _2^2,
\end{aligned}
\end{equation}
where $F_{ref}$ denotes the corresponding flare reference and $\nabla$ is the gradient operation.

During each restoration step, the mask generation network (MGN) with an architecture of two convolutional layers takes $1 - F_k$ as input and outputs the spatial attention mask $M_k$. Then the restoration network with a Uformer \cite{Uformer} architecture takes $M_k \odot I_k$ as input and produces recovered result $\hat{I}_{k+1}$, where $\odot$ is element-wise multiplication. The reconstruction loss is as follows:
\begin{align} 
\begin{split}
\mathcal{L}_{re} &= \sum_{k=1}^K \gamma_k ( \Vert \hat{I}_k - I_{ref}\Vert _F^2  +  1 - SSIM(\hat{I}_k - I_{ref}))\\
&+ \Vert\phi(\hat{I_k}) - \phi(I_{ref}) \Vert _1,
\end{split}
\end{align}
where $K$ is the total number of recurrent times set to be 3, $\phi(\cdot)$ represents a pre-trained VGG19 feature extractor, $SSIM(\cdot)$ denotes the structural similarity loss, and $\gamma_k$ is hyper-parameter set to 1/32, 1/8, and 1 in different steps.
Besides, those severely polluted pixels that are nearer the brightest light source should be assigned small weights, such that the generated attention mask is similar to $1 - F_k$, which leads to a mask loss:
\begin{equation}
\mathcal{L}_{m} = \lambda \sum_{k=1}^K \Vert \hat{M}_k - (1 - F_k)\Vert _F^2,
\end{equation}
where $\lambda$ is a hyper-parameter set to be 0.1. The total loss function is $\mathcal{L}_{total}=\mathcal{L}_{re}+\mathcal{L}_{m}$.

The authors train FDN for 300 epochs alone, and train the MGN and Uformer for 200 epochs together with batch size and learning rate set to be 4 and 0.0001 respectively. All the experiments are implemented on Nvidia A100 GPU with pytorch framework.

\begin{figure*}[t]
	\centering
	\includegraphics[width=0.6\linewidth]{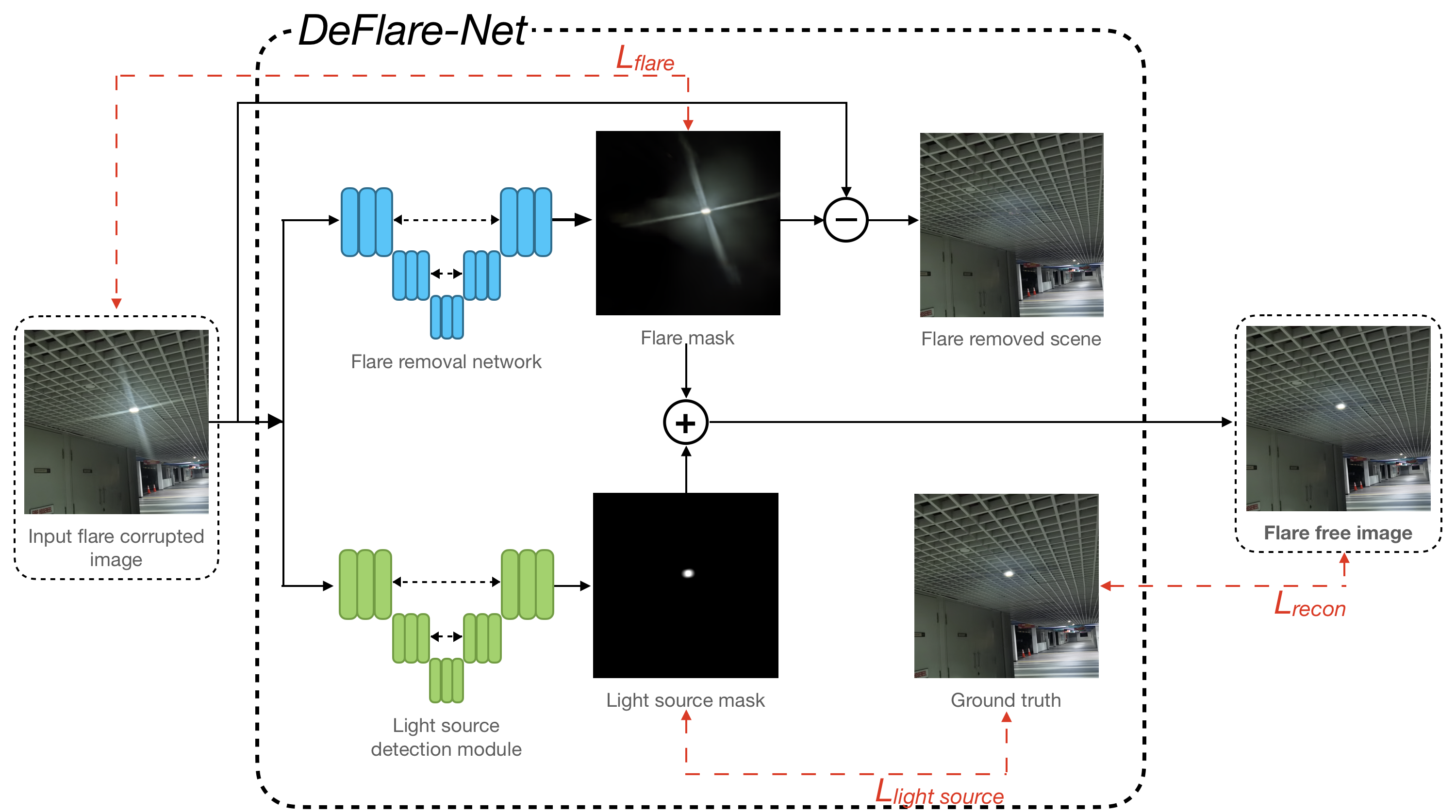}
	\caption{The network architecture of DeFlare-Net for removal of flare artefacts proposed by CEVI\_Explorers team.}
    \vspace{-4mm}
	\label{fig:DFN}
\end{figure*}

\paragraph{\bf CEVI\_Explorers.}This team proposes a pipeline DeFlare-Net for the removal of flare artifacts from images. The proposed architecture (DeFlare-Net) is shown in Figure \ref{fig:DFN}. DeFlare-Net comprises two main modules, i.e., Flare Removal Network (FRN), and Light Source Detection Network (LSDN) as shown in Figure \ref{fig:DFN}. FRN and LSDN are built upon U-Net-based architecture \cite{wong} with skip connection across scales. Given a flared image input $I_{flare}$, the aim of the proposed DeFlare-Net is to learn flare artifacts, and light source and generate a flare-free image. FRN detects and removes flare by masking the flare region $X_{flare}$. FRN implicitly learns to segment light source along with flare region, as shown in Figure \ref{fig:DFN}. Removal of the light source creates additional artifacts resulting in unpleasant observation. Towards this, the authors introduce LSDN to retain the light source in accordance with the ground truth image/true scene. LSDN learns to segment the light source from the scene for retaining the light source in the flare-free image. The removal of flare from the input flare image $I_{flare}$ is guided by the flare mask, resulting in the implicit removal of a light source. To retain the light source in the image, the light source mask is blended with the flare-removed scene, to obtain the final flare-free image.

{\bf Loss Functions.} To optimize the task of flare removal using DeFlare-Net the authors propose $L_{DeFlare}$ as a weighted combination of Flare loss $L_{flare}$, Lightsource loss $L_{ls}$, and Reconstruction loss $L_{recon}$.
To learn the segmentation of flares in the images, we incorporate $L_{flare}$ \cite{unpaired}. $L_{flare}$ is computed between the input flare image and the flare mask. 
To learn the segmentation of light source in the true scene/ground truth image, the authors incorporate light source loss $L_{ls}$. $L_{ls}$ is computed between the light source masks of the ground truth image and the predicted mask. To facilitate the overall restoration of the flare-free image the authors use $L_{recon}$ computed between ground truth image and the flare-free image. The proposed $L_{DeFlare}$ loss is given as,
\begin{equation}
  \mathbb{L}_{DeFlare} = \alpha * \mathbb{L}_{flare} + \beta*\mathbb{L}_{ls} + \gamma*\mathbb{L}_{recon},
\end{equation}
where $\alpha$, $\beta$, and $\gamma$ are weights and we heuristically set $\alpha=\beta=\gamma=0.33$.

The DeFlare-Net architecture was written using Python (v3.8) and PyTorch framework. All experiments were performed on NVIDIA RTX 3090 GPU with 24GB memory and AMD Ryzen Threadripper processor.

\begin{figure*}[t]
	\centering
	\includegraphics[width=0.7\linewidth]{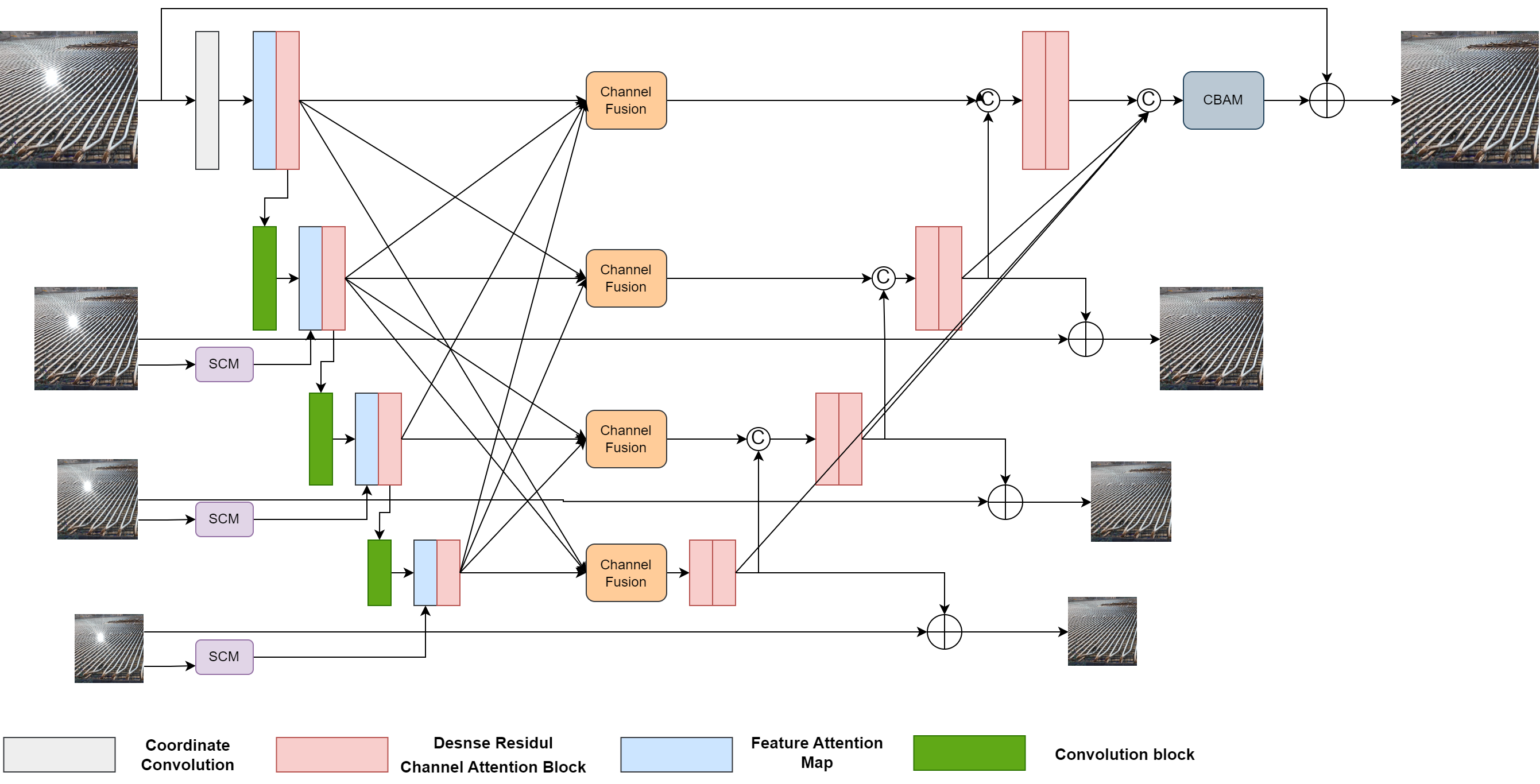}
	\caption{The network architecture of Couger AI team.}
 \vspace{-5mm}
	\label{pipeline_cougerAI}
\end{figure*}

\paragraph{\bf AI ISP.}
This team adopts Uformer \cite{Uformer} architecture and makes several improvements in loss function and data synthesis to boost the network’s performance.

Using L1 loss and focal frequency loss, the flare regions should be given a higher weight when calculating the loss function to get better restoration results in these regions. Thus, both global loss like \cite{how_to} and regional loss are used. The predicted image is replaced by $\hat{I}_{g}$ and $\hat{I}_{r}$ when calculating the global loss and regional loss respectively:
\begin{align}
\begin{split}   
&\hat{I}_{g} = I_{0}\odot M_l  + f(I_f,\Theta)\odot(1-M_l),\\
&\hat{I}_{r} = I_{0}\odot (M_l \vee M_{nf}) + f(I_f,\Theta)\odot(1- (M_l \vee M_{nf})),
\end{split}
\end{align}
where $M_l$ and $M_{nf}$  denote binary masks indicating the light source and non-flare regions respectively, $I_0$ and $I_f$ denote the clean image and flare-polluted image respectively, $f(I_f,\Theta)$denotes the flare removal network, $\odot$ denotes Hadamard product, and $\vee$ denotes logic OR. The masks are generated using threshold segmentation when loading the training samples. The total loss function is the weighted sum of the global loss and regional loss.

To get better performance on  nighttime scenes, this team uses extra 260 RAW2RGB images from SID \cite{chenLearningSeeDark2018}, which are manually cropped  to remove  flare and light source regions. The model is trained for 80 epochs on the Flickr24K \cite{flickr24K} and SID dataset, and for another 60 epochs on Flickr24K only. Besides, in view of the fact that there may exist more than one light source in real images, this team synthesizes images with multi light sources of different scales in addition to images with only a single light source.


The proposed solution is implemented based on PyTorch  on Nvidia A100 GPUs. Each batch contains 16 samples, among which 11 images have one light source, four images have two light sources, and one image has three light sources. Adam optimizer is applied with learning rate $10^{-4}$.

\paragraph{\bf Couger AI.}
The network is inspired by \cite{feng2023mipi,nathan2021skeletonnetv2}. The proposed network is a multi-level U-net-based system (see Figure~\ref{pipeline_cougerAI}). It includes four encoder and four decoder blocks. The input image is passed to the Coordinate\cite{liu2018intriguing} convolution layer to enhance the spatial quality. The output of the Coordinate convolution layer passed to the Encoder block. The encoder block contains the convolution layer with a 3x3 filter, Feature Attention map block, and  Dense Residual Channel Attention Block\cite{zhang2018residual} [DRCA], followed by a downsampling layer. The output of the encoder block is passed to the channel fusion block and concatenated to the corresponding decoder block. The decoder block contains the upsampling layer with two DRCA blocks. Each decoder block is fused with the input image and supervised by the loss function. Each decoder level's output is concatenated along the channel axis, followed by the CBAM\cite{woo2018cbam} attention block before the final supervision.

\section{Acknowledgements}
We thank Nanyang Technological University, Shanghai Artificial Intelligence Laboratory, Sony and The Chinese University of Hong Kong to sponsor this MIPI 2023 challenge. We thank all the organizers for their contributions to this workshop and all the participants for their great work.

{\small
\bibliographystyle{ieee_fullname}
\bibliography{egbib}
}

\appendix

\section{Teams and Affiliations}
\label{append:teams}

\subsection*{\bf MiAlgo.}
\noindent
{\bf Title:} Erasing and Inpainting network for nighttime flare removal\\
{\bf Members:}\\
Shuai Liu$^1$ (\href{liushuai21@xiaomi.com}{liushuai21@xiaomi.com})\\
Hao Wang$^1$\quad Chaoyu Feng$^1$\quad Luyang Wang$^1$\quad Guangqi Shao$^1$\quad Chenguang Zhang$^1$\quad Xiaotao Wang$^1$\quad Lei Lei$^1$\\
{\bf Affiliations:}\\
$^1$ Xiaomi Inc., China\\

\subsection*{\bf SRC-B.}
\noindent
{\bf Members:}\\
Dafeng Zhang$^1$ (\href{dfeng.zhang@samsung.com}{dfeng.zhang@samsung.com})\\
Xiangyu Kong$^1$\quad Guanqun Liu$^1$\quad Mengmeng Bai$^1$\quad Jia Ouyang$^1$\quad Xiaobing Wang$^1$\quad Jiahui Yuan$^1$\\
{\bf Affiliations:}\\
$^1$ Samsung Research China - Beijing (SRC-B), China\\

\subsection*{\bf MegFR.}
\noindent
{\bf Title:} Nighttime Flare Removal through Signal Separation\\
{\bf Members:}\\
Xinpeng Li$^1$ (\href{lixinpeng@megvii.com}{lixinpeng@megvii.com})\\
Chengzhi Jiang$^1$, Ting Jiang$^1$, Wenjie Lin$^1$, Qi Wu$^1$, Mingyan Han$^1$, Jinting Luo$^1$, Lei Yu$^1$, Haoqiang Fan$^1$ and Shuaicheng Liu$^{2,1*}$\\
{\bf Affiliations:}\\
$^1$ Megvii Technology\\
$^2$ University of Electronic Science and Technology of
China (UESTC)\\

\subsection*{\bf AntIns.}
\noindent
{\bf Title:} An Effective Nighttime Flare Removal Pipeline for MIPI Challenge\\
{\bf Members:}\\
Bo Yan (\href{lengyu.yb@antgroup.com}{lengyu.yb@antgroup.com})\\
Zhuang Li \quad Yadong Li \quad Hongbin Wang \\
{\bf Affiliations:}\\
Ant Group, China\\

\subsection*{\bf ActionBrain-ETRI.}
\noindent
{\bf Title:} Suppressing Flares in the Dark Images with
Cascaded U-Former Architecture\\
{\bf Members:}\\
Soonyong Song  (\href{soony@etri.re.kr}{soony@etri.re.kr})\\
{\bf Affiliations:}\\
Electronics and Telecommunications Research Institute (ETRI), Daejeon, South Korea\\

\subsection*{\bf USask-Flare.}
\noindent
{\bf Title:} A Light source guided Spatial transformer Generative Adversarial Network for single Image Flare Removal\\
{\bf Members:}\\
Minghan Fu (\href{fig072@mail.usask.ca}{fig072@mail.usask.ca})\\
Rayyan Azam Khan \quad Fangxiang Wu \\
{\bf Affiliations:}\\
College of Engineering, University of Saskatchewan, Canada\\

\subsection*{\bf LVGroup\_HFUT.}
\noindent
{\bf Title:} Flare-guide nonlinear activation free network for nighttime flare removal\\
{\bf Members:}\\
Zhao Zhang$^1$ (\href{cszzhang@gmail.com}{cszzhang@gmail.com})\\
Suiyi Zhao$^1$\quad Huan Zheng$^1$\quad Yangcheng Gao$^1$\quad Yanyan Wei$^1$\quad Jiahuan Ren$^1$\quad Bo Wang$^1$\quad Yan Luo$^1$\\
{\bf Affiliations:}\\
$^1$ Laboratory of Multimedia Computing, Hefei University of Technology, China\\

\subsection*{\bf szzzzz01.}
\noindent
{\bf Title:} A Flare-region Attention Recurrent Network for Flare Removal\\
{\bf Members:} \\
Shuaibo Gao$^1$ (\href{gaosb201@gmail.com}{gaosb201@gmail.com})\\ 
Wenhui Wu$^1$ \quad Sicong Kang$^1$\\
{\bf Affliations:}\\
College of Electronics and Information Engineering, Shenzhen University, China\\

\subsection*{\bf CEVI\_Explorers.}
\noindent
{\bf Title:} DeFlare-Net: Flare Detection and Removal Network\\
{\bf Members:} \\
Nikhil Akalwadi (\href{nikhil.akalwadi@kletech.ac.in}{nikhil.akalwadi@kletech.ac.in})\\ 
Ankit Raichur \quad Vinod Patil \quad Allabakash G\\ 
Swaroop A \quad Amogh Joshi \quad Chaitra Desai\\ 
Ramesh Ashok Tabib \quad Ujwala Patil \quad Uma Mudenagudi\\
{\bf Affliations:}\\
Center of Excellence in Visual Intelligence (CEVI)- KLE Technological University, Hubballi, Karnataka, India\\

\subsection*{\bf AI ISP.}
\noindent
{\bf Title:} Improvements of loss function and data synthesis for nighttime flare removal\\
{\bf Members:}\\ 
Ruoxi Zhu$^1$\quad Jiazheng Lian$^2$(\href{jzlian20@fudan.edu.cn}{jzlian20@fudan.edu.cn})\\
 Sicheng Li \quad Shusong Xu \quad Zihao Liu\\
{\bf Affiliations:} \\
$^1$School of Microelectronics, Fudan University, China\\
$^2$Academy for Engineering and Technology, Fudan University, China\\

\subsection*{\bf Couger AI.}
\noindent
{\bf Title:} Light Weight Dense Residual Channel attention model for flare removal\\
{\bf Members:}\\
Sabari Nathan$^1$ (\href{sabari@couger.co.jp}{sabari@couger.co.jp})\\
Priya Kansal$^1$\\
{\bf Affiliations:}\\
$^1$ Couger Inc, Tokyo, Japan\\

\end{document}